\documentclass[runningheads]{llncs}
\usepackage{multirow}

\usepackage[mathscr]{eucal} 
\usepackage{amsbsy} 
\usepackage{times}
\usepackage{latexsym}
\usepackage{float} 
\usepackage{cite}
\usepackage{amsfonts}
\usepackage[table]{xcolor}  
\usepackage{array}          
\usepackage[T1]{fontenc}
\usepackage{wrapfig}
\usepackage{booktabs}
\usepackage{xcolor}
\usepackage{subcaption}
\usepackage[colorlinks=true,linkcolor=black,citecolor=darkblue,urlcolor=darkblue]{hyperref} 

\newcommand{\mymethod}{ACTGNN}

\definecolor{darkblue}{RGB}{0,0,139}

\usepackage[utf8]{inputenc}

\usepackage{inconsolata}

\usepackage{graphicx}
\usepackage{amsmath}
\begin{document}
%
\title{ACTGNN: Assessment of Clustering Tendency with Synthetically-Trained Graph Neural Networks}
%
%
\author{First Author\inst{1}\orcidID{0000-1111-2222-3333} \and
Second Author\inst{2,3}\orcidID{1111-2222-3333-4444} \and
Third Author\inst{3}\orcidID{2222--3333-4444-5555}}

\author{Yiran Luo, Evangelos E. Papalexakis}
\institute{University of California Riverside, Riverside, CA, USA\\ \email{yluo147@ucr.edu, epapalex@cs.ucr.edu}}


\maketitle              
\begin{abstract}
Determining clustering tendency in datasets is a fundamental but challenging task, especially in noisy or high-dimensional settings where traditional methods, such as the Hopkins Statistic and Visual Assessment of Tendency (VAT), often struggle to produce reliable results. In this paper, we propose \mymethod{}, a graph-based framework designed to assess clustering tendency by leveraging graph representations of data. Node features are constructed using Locality-Sensitive Hashing (LSH), which captures local neighborhood information, while edge features incorporate multiple similarity metrics, such as the Radial Basis Function (RBF) kernel, to model pairwise relationships. A Graph Neural Network (GNN) is trained exclusively on synthetic datasets, enabling robust learning of clustering structures under controlled conditions. Extensive experiments demonstrate that \mymethod{} significantly outperforms baseline methods on both synthetic and real-world datasets, exhibiting superior performance in detecting faint clustering structures, even in high-dimensional or noisy data. Our results highlight the generalizability and effectiveness of the proposed approach, making it a promising tool for robust clustering tendency assessment.

\keywords{Clustering Tendency  \and Graph Neural Networks \and Synthetic Data \and Locality-Sensitive Hashing}
\end{abstract}

\section{Introduction}
\label{sec:intro}

Clustering is a fundamental task in data analysis, crucial for uncovering hidden patterns and structures within complex datasets. Its applications span diverse fields, from fraud detection in banking systems to hierarchical pixel clustering for image segmentation~\cite{ghosal2020short}. Most clustering algorithms require a predefined number of clusters as input. However, providing an inaccurate number can lead to suboptimal or misleading clustering results. Consequently, determining whether a dataset contains an underlying cluster structure---and subsequently identifying the number of clusters---is a critical step. This process, referred to as the \textit{assessment of clustering tendency}, begins with the fundamental question: does the dataset exhibit clustering structures at all?

Assessing clustering tendency presents significant challenges. High dimensionality, common in modern datasets, can obscure meaningful clusters by introducing irrelevant features. Similarly, noise and outliers may distort the data, creating false cluster-like structures or masking true ones. For instance, in image segmentation, noise may introduce false edges that mimic clusters, while in high-dimensional gene expression data, irrelevant features often obscure meaningful patterns.

To address these challenges, several methods have been proposed. The Hopkins Statistic~\cite{hopkins1954new} is a popular statistical test that estimates whether a dataset resembles a uniform random distribution. Another widely used method, VAT (Visual Assessment of Tendency)~\cite{bezdek2002vat}, generates visual representations to aid in clustering assessment. Although effective in specific scenarios, these methods have limitations. The Hopkins Statistic is sensitive to sample size and outliers, reducing its reliability on noisy and higher-dimensional datasets. VAT, on the other hand, relies heavily on subjective visual inspection, which can be ambiguous, particularly for complex or high-dimensional data. Given these limitations, there is a pressing need for robust and automated methods to assess the clustering tendency, particularly in complex datasets.

Recent advancements in graph neural networks (GNNs) have shown promise in clustering tasks. For instance, Tsitsulin et al.~\cite{tsitsulin2023graph} introduced Deep Modularity Networks (DMoN), an unsupervised GNN pooling method inspired by modularity-based clustering quality, demonstrating significant improvements in graph clustering. Similarly, Bhowmick et al.~\cite{bhowmick2024dgcluster} proposed DGCluster, a framework that optimizes the modularity objective using GNNs, achieving state-of-the-art results in attributed graph clustering. These developments highlight the potential of GNN-based approaches in clustering applications, motivating their use in clustering tendency assessment.

\begin{figure}[ht]
    \centering
    \includegraphics[width=0.99\textwidth]{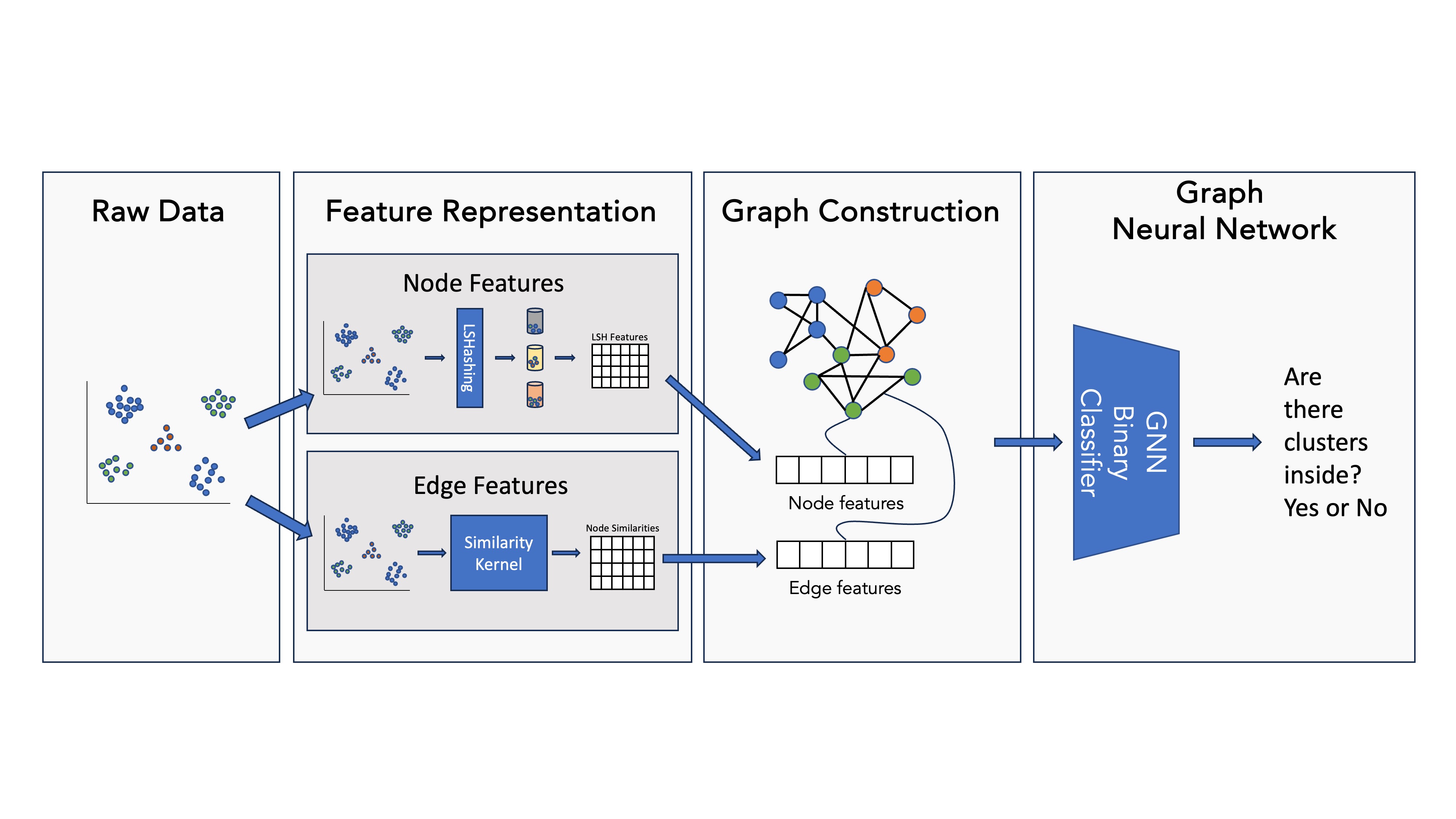}
    \caption{Overview of the proposed \mymethod{} framework for clustering tendency assessment. The process includes transforming raw data into a graph representation by constructing node and edge features, followed by binary classification using a graph neural network.}
    \label{fig:framework_overview}
\end{figure}

In this paper, we propose Assessment of Clustering Tendency with Synthetically-Trained Graph Neural Networks (\mymethod{}). Figure~\ref{fig:framework_overview} illustrates the overall pipeline of our proposed method. The framework begins with raw data, which is transformed into a graph representation through carefully designed node and edge features. The resulting graph is then processed by a GNN to determine whether a k-means clustering structure exists in the dataset. GNNs are well-suited for this task due to their ability to model pairwise relationships as graphs, enabling robust identification of clustering structures even in noisy, high-dimensional data. A notable feature of our approach is that the model is trained only on synthetic data but evaluated on both synthetic and real-world datasets. The use of synthetic data during training ensures control over clustering structures and noise, allowing us to systematically evaluate the model’s generalizability across diverse test scenarios.

Specifically, our contributions are as follows:
\begin{itemize}
    \item \textbf{Graph-Based Framework Design}: We propose a novel framework that represents datasets as graphs with carefully designed node and edge features, enabling efficient detection of clustering structures.
    
    \item \textbf{Comprehensive Node and Edge Feature Construction}: We introduce robust node and edge feature strategies to enhance the expressiveness of the graph representation:
    \begin{itemize}
        \item \textbf{Node Features}: We employ Locality-Sensitive Hashing (LSH) to construct node features, capturing local structural information by summarizing distances to neighboring nodes.
        \item \textbf{Edge Features}: Multiple edge feature options are explored, including unweighted edges, Euclidean distance, cosine similarity, and the Radial Basis Function (RBF) kernel for measuring similarity between connected nodes.
    \end{itemize}
    
    \item \textbf{Synthetic-to-Real Generalization}: We train our graph neural network (GNN) model exclusively on synthetic data and demonstrate its strong generalization capabilities by evaluating it on both synthetic and real-world datasets, achieving significant improvements in clustering tendency assessment over traditional methods.
\end{itemize}


All code from this study will be publicly available upon publication to support transparency and future research.\footnote{https://anonymous.4open.science/r/ACTGNN-3F24/}

\section{Related Work}
\label{sec:related}


Assessing clustering tendency is a fundamental step in data analysis, determining the presence of inherent groupings within datasets. Traditional methods like the Hopkins Statistic~\cite{hopkins1954new} and Visual Assessment of Tendency (VAT)~\cite{bezdek2002vat} have been widely utilized. However, these approaches often face challenges with high-dimensional data and noise, prompting the exploration of advanced techniques. This section reviews relevant work in three key areas: traditional clustering tendency assessment, Graph Neural Networks (GNNs) for clustering, and the use of synthetic data in machine learning models.

\subsection{Traditional Methods for Clustering Tendency Assessment}
Traditional approaches, such as the Hopkins Statistic and VAT, have been extensively used for assessing clustering tendency. The Hopkins Statistic is a statistical measure that tests the spatial randomness of a dataset, with higher values indicating clustering structures~\cite{hopkins1954new}. VAT, on the other hand, provides a visual representation of pairwise dissimilarity matrices to reveal clustering structures~\cite{bezdek2002vat}.  

To address the limitations of VAT, particularly its reliance on subjective visual interpretation, several automated variants have been developed. The improved VAT (iVAT) algorithm enhances VAT's effectiveness by applying a path-based distance transform, enabling better performance in complex datasets~\cite{havens2011efficient}. Automated VAT (aVAT) integrates cluster detection mechanisms into the VAT framework, reducing subjectivity and improving automation~\cite{wang2010ivat}. Recently, HaVAT extends these efforts by providing an automatic assessment of cluster structures in unlabeled data, offering further improvements in robustness and accuracy~\cite{pagadala2024havat}.

\subsection{Graph Neural Networks for Clustering}
Graph Neural Networks (GNNs) have emerged as powerful tools for learning from graph-structured data, achieving state-of-the-art results in tasks such as node classification and link prediction. Their application in clustering tasks has demonstrated significant potential for improving clustering quality and robustness.

For example, Tsitsulin et al. introduced Deep Modularity Networks (DMoN), an unsupervised GNN pooling method that leverages modularity measures to improve graph clustering~\cite{tsitsulin2023graph}. Similarly, Bhowmick et al. proposed DGCluster, a GNN-based approach that optimizes the modularity objective, achieving state-of-the-art performance on attributed graph clustering tasks~\cite{bhowmick2024dgcluster}. These studies demonstrate the suitability of GNN-based methods for learning complex relationships and structural patterns in clustering problems.

\subsection{Synthetic Data for Model Training}
The use of synthetic data for training machine learning models has gained significant traction due to its benefits, such as dataset augmentation, privacy preservation, and the ability to create controlled evaluation scenarios. Yuan et al. analyzed the principles of training data synthesis for supervised learning, proposing a framework to optimize synthesis efficacy from a distribution-matching perspective~\cite{yuan2023real}. In graph learning, Tsitsulin et al. explored synthetic graph generation to benchmark graph learning algorithms, providing a foundation for controlled experimentation~\cite{tsitsulin2022synthetic}.  

In the context of clustering, Zhang et al. introduced the AnchorGAE model, which utilizes synthetic data to enhance clustering performance through efficient bipartite graph convolution~\cite{zhang2022non}. Additionally, models like Frappe~\cite{shiao2024frappe} further demonstrate the utility of synthetic data in improving model performance under controlled conditions. However, challenges such as ensuring the diversity and representativeness of synthetic data remain, as poor-quality synthetic data can degrade model generalization to real-world datasets.

\section{Methodology}
\label{sec:method}

We propose \mymethod{}, a learning-based framework to determine whether a given dataset exhibits a k-means clustering structure. To construct the graph representation, we treat each data point as a node and connect it to its \( K \)-nearest neighbors (KNN), forming a graph that encodes local relationships between data points. This graph serves as input to a Graph Neural Network (GNN), which performs the binary classification task.

Although the KNN graph provides the structural backbone, the features of the nodes and edges play a critical role in capturing underlying patterns. Below, we describe the construction of node and edge features, followed by the GNN design.

\subsection{Node Features}
We construct node features using Locality-Sensitive Hashing (LSH), a method that efficiently captures the local neighborhood properties of each data point in high-dimensional space. LSH allows us to approximate nearest neighbors through hash-based indexing, providing a compact and informative representation of the relationships between points.

Locality-Sensitive Hashing works by mapping high-dimensional data points into lower-dimensional buckets using a series of hash functions that preserve proximity. Each data point is indexed into multiple hash tables, where each table applies a random hash function to assign the point to a specific bucket. Using multiple hash tables increases robustness, ensuring that similar points are more likely to be hashed into the same bucket while reducing false negatives.

Once the points are indexed, we query the nearest neighbors of each data point within the hashed buckets. The number of neighbors is dynamically set as a percentage of the dataset size, capped to ensure computational efficiency. The neighbors are identified using Euclidean distance within the buckets.

The local neighborhood structure for each node is summarized by aggregating distances to its nearest neighbors. For every data point, the following features are computed:
\begin{itemize}
    \item The average Euclidean distance to the nearest neighbors, providing an estimate of the node's local proximity.
    \item The number of neighbors returned by the LSH query, which reflects the density of points in the neighborhood.
    \item The variance of the distances, capturing the spread or variability within the local neighborhood.
\end{itemize}



\subsection{Edge Features}
Edges in the graph encode relationships between nodes, providing critical information about local and global structural patterns. We construct edges using a \( K \)-nearest neighbors (KNN) approach, where each node is connected to its \( K \)-closest neighbors based on a chosen similarity metric. The edge features are then derived from these relationships, allowing the model to distinguish between connected nodes based on their proximity or similarity. We consider four types of edge features:

\paragraph{1. Unweighted Edges}  
In this approach, edges are treated as unweighted, capturing only the graph's connectivity structure without assigning explicit features. 

\paragraph{2. Euclidean Distance}  
The Euclidean distance between connected nodes is used as an edge feature. Given two nodes \( i \) and \( j \) with positions \( x_i \) and \( x_j \), the edge weight is defined as:
\[
e_{ij} = \|x_i - x_j\|_2,
\]
where \( \| \cdot \|_2 \) is the \( \ell_2 \)-norm. 

\paragraph{3. Cosine Similarity}  
Cosine similarity measures the angular similarity between the feature vectors of connected nodes. For two nodes \( i \) and \( j \), the edge weight is calculated as:
\[
e_{ij} = \frac{x_i \cdot x_j}{\|x_i\| \|x_j\|},
\]
where \( x_i \cdot x_j \) is the dot product of the node features, and \( \|x_i\| \) and \( \|x_j\| \) are their magnitudes. 

\paragraph{4. Radial Basis Function (RBF) Kernel}  
The RBF kernel provides a non-linear measure of similarity between nodes based on their pairwise distance. For two nodes \( i \) and \( j \), the edge weight is defined as:
\[
e_{ij} = \exp\left(-\frac{\|x_i - x_j\|^2}{2\sigma^2}\right),
\]
where \( \sigma \) is a scaling parameter that controls the influence of distance. 

To determine the optimal edge feature strategy and the percentage of nearest neighbors connected, we conducted a grid search over various configurations (e.g., unweighted, Euclidean distance, cosine similarity, and RBF kernel with different $\sigma$ values). The results, detailed in the appendix, guided our choice of edge features and connectivity parameters.

\subsection{Graph Neural Network}
Once the graph is constructed with the defined node and edge features, it is fed into a Graph Neural Network (GNN) for binary classification. GNNs aggregate and learn structural relationships by iteratively passing information between nodes and edges, making them well-suited for identifying clustering structures.

We employ a Graph Convolutional Network (GCN), a widely-used GNN variant. The GCN updates each node’s representation by aggregating features from its neighbors, capturing both local and global structural information. Our framework uses a five-layer GCN, followed by a global mean pooling layer and a fully connected layer for classification.

The output of the GCN is a binary prediction indicating whether the dataset exhibits a k-means clustering structure, leveraging the node and edge features designed in \mymethod{}.

\section{Experiments and Results}
\label{sec:experiments}

We evaluate the proposed \mymethod{} on both synthetic and real-world datasets to comprehensively assess its performance in detecting clustering structures. All experiments in this paper use the Radial Basis Function (RBF) kernel with $\sigma = 2$ as the edge feature strategy, and 60\% of the nearest neighbors are connected based on the results of our grid search analysis (see Appendix).

\subsection{Synthetic Test Datasets}
To evaluate the proposed \mymethod{}, we compare its performance against two baseline methods designed around the principles of the Hopkins Statistic~\cite{hopkins1954new} and the K-means with Silhouette Score~\cite{macqueen1967some, lloyd1982least, rousseeuw1987silhouettes}. Specifically, the first baseline uses the Hopkins Statistic as a measure of clustering tendency, while the second baseline employs a threshold-based approach using the Silhouette Score computed from K-means clustering results. All experiments are conducted on synthetic datasets, with the \mymethod{} model trained exclusively on synthetic data.

\paragraph{First Baseline: Hopkins-Statistic-Based Method}  
The first baseline is based on the \textbf{Hopkins Statistic}, which measures clustering tendency by comparing the distribution of points to a uniform random distribution. The Hopkins score ranges between 0 and 1, where values above 0.75 typically indicate significant clustering structures. However, no universally accepted threshold exists, leading to ambiguity in its direct application. To address this, we design a threshold-based method that classifies datasets as clustered or non-clustered based on a range of predefined thresholds from 0.6 to 0.9, incremented by 0.05. Lower thresholds are more permissive, potentially identifying weak clustering structures, while higher thresholds are stricter but may miss moderate clustering.

\paragraph{Second Baseline: Silhouette-Score-Based Method}  
The second baseline is based on the \textbf{K-means} clustering algorithm and the \textbf{Silhouette Score}, which evaluates the quality of clustering results. The Silhouette Score measures how similar a point is to its assigned cluster relative to other clusters, ranging from -1 to 1, where higher values indicate well-separated and cohesive clusters. In our method, we apply K-means clustering over a range of \( k \)-values, starting from 2 and capped at a maximum of 20, depending on the dataset size. For each dataset, the maximum Silhouette Score obtained across the range of \( k \)-values is compared to a predefined threshold to determine clustering tendency. Since no universally accepted threshold exists, we test multiple values between 0.3 and 0.75, incremented by 0.05. Lower thresholds detect weaker clustering structures, while higher thresholds are stricter and may overlook datasets with moderate clustering.


\begin{figure}[ht]
    \centering
    \begin{subfigure}[b]{0.48\textwidth}
        \centering
        \includegraphics[width=\textwidth]{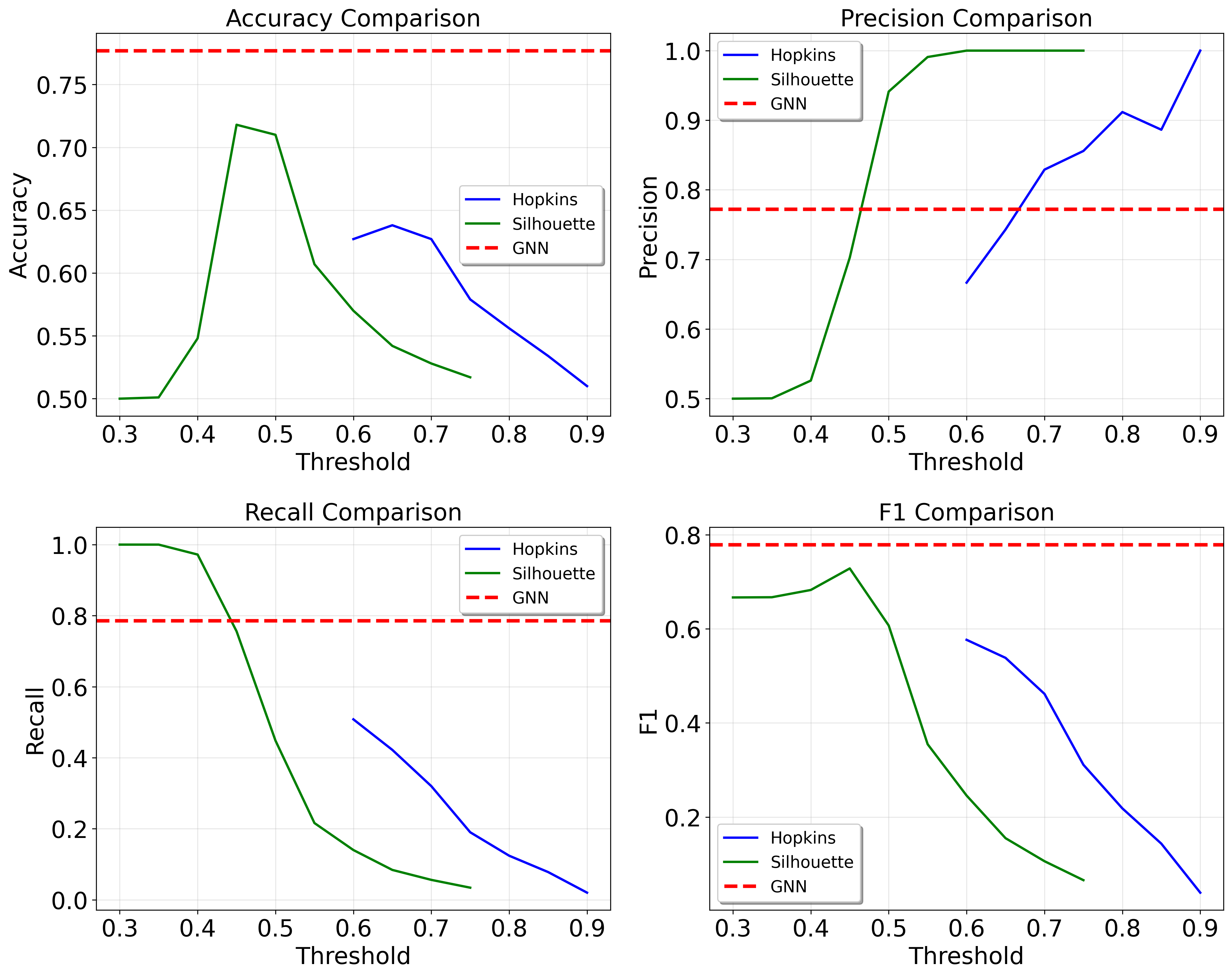}
        \caption{Performance comparison on 2D datasets.}
        \label{fig:comparison_2d}
    \end{subfigure}
    \hfill
    \begin{subfigure}[b]{0.48\textwidth}
        \centering
        \includegraphics[width=\textwidth]{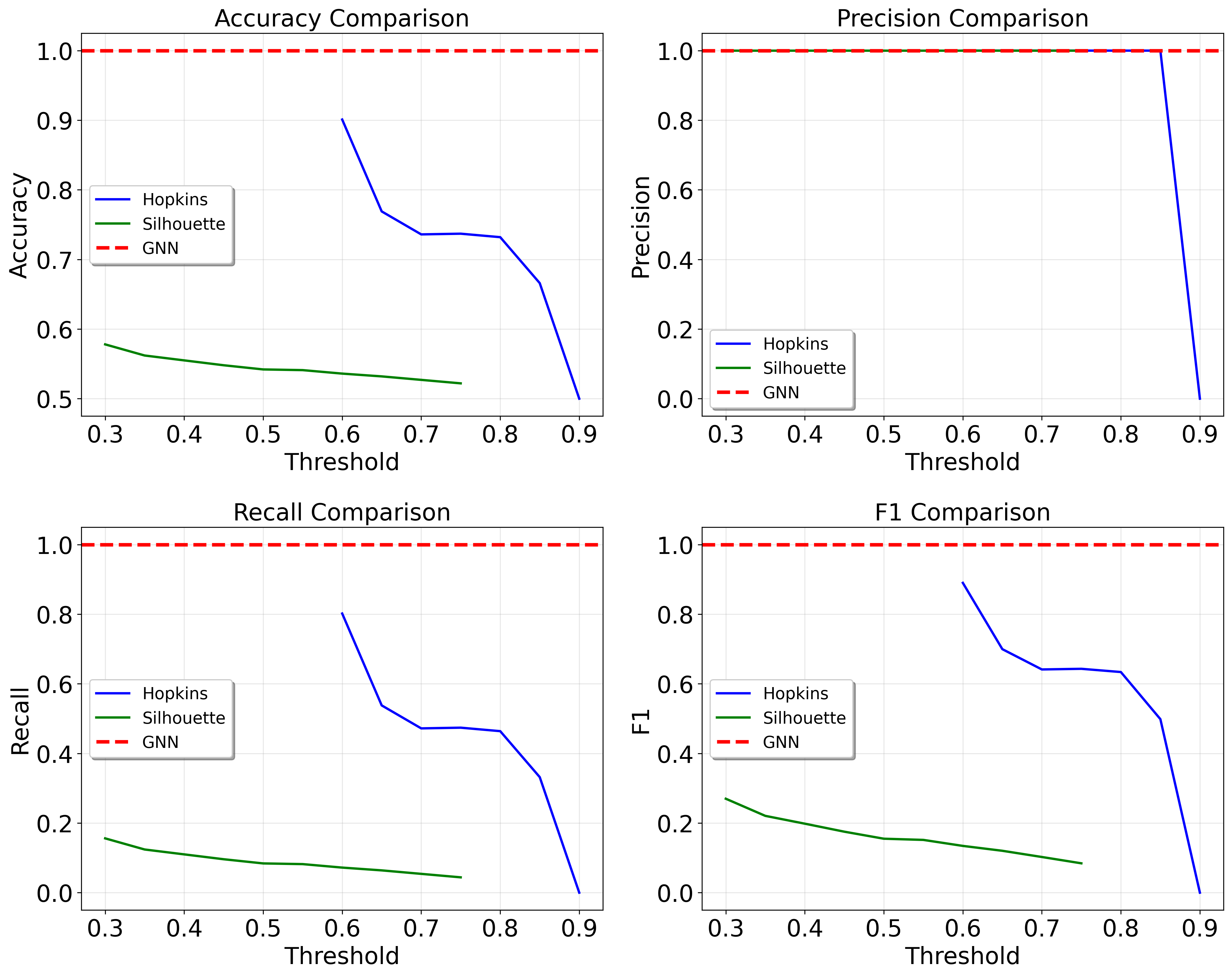}
        \caption{Performance comparison on 30D datasets.}
        \label{fig:comparison_30d}
    \end{subfigure}
    \caption{Performance comparison of the \mymethod{}, Hopkins Statistic, and K-means with Silhouette Score on synthetic datasets of different dimensions. The horizontal red dashed line represents the \mymethod{}'s performance.}
    \label{fig:method_comparison}
\end{figure}



In Figure~\ref{fig:method_comparison}, we compare the performance of \mymethod{}, Hopkins Statistic, and K-means with Silhouette Score on synthetic datasets for two dimensions: 2D (Figure~\ref{fig:comparison_2d}) and 30D (Figure~\ref{fig:comparison_30d}). The evaluation metrics—accuracy, precision, recall, and F1 score—are plotted across varying thresholds for the two baseline methods, with \mymethod{}'s performance shown as a horizontal red dashed line.

In the 2D case (Figure~\ref{fig:comparison_2d}), the Silhouette Score peaks around mid-range thresholds (0.5–0.6) but declines as thresholds tighten, while the Hopkins Statistic struggles to maintain accuracy and recall. \mymethod{} consistently outperforms both baselines with stable, high scores. In the 30D case (Figure~\ref{fig:comparison_30d}), baseline performances degrade significantly, particularly at higher thresholds. However, \mymethod{} maintains near-perfect results, highlighting its robustness in high-dimensional settings.

\begin{table*}[ht]
    \centering
    \caption{Performance comparison across varying dimensions. For each dimension, the best accuracy and F1 score are shown in \textbf{bold}.}
    \label{tab:dimension_comparison}
    \begin{tabular}{|c|c|c|c|c|}
        \hline
        \textbf{Dimension} & \textbf{Method} & \textbf{Accuracy} & \textbf{F1 Score} & \textbf{Precision / Recall} \\
        \hline
        \multirow{3}{*}{2} 
        & GNN & \textbf{0.777} & \textbf{0.779} & 0.772 / 0.786 \\
        & Hopkins & 0.627 & 0.577 & 0.667 / 0.508 \\
        & Silhouette & 0.718 & 0.728 & 0.703 / 0.756 \\
        \hline
        \multirow{3}{*}{3}
        & GNN & \textbf{0.838} & \textbf{0.853} & 0.783 / 0.936 \\
        & Hopkins & 0.717 & 0.677 & 0.788 / 0.594 \\
        & Silhouette & 0.820 & 0.809 & 0.860 / 0.764 \\
        \hline
        \multirow{3}{*}{5}
        & GNN & \textbf{0.957} & \textbf{0.957} & 0.956 / 0.958 \\
        & Hopkins & 0.813 & 0.791 & 0.898 / 0.706 \\
        & Silhouette & 0.701 & 0.574 & 1.000 / 0.402 \\
        \hline
        \multirow{3}{*}{10}
        & GNN & \textbf{0.991} & \textbf{0.991} & 1.000 / 0.982 \\
        & Hopkins & 0.821 & 0.787 & 0.974 / 0.660 \\
        & Silhouette & 0.585 & 0.291 & 1.000 / 0.170 \\
        \hline
        \multirow{3}{*}{20}
        & GNN & \textbf{0.996} & \textbf{0.996} & 0.992 / 1.000 \\
        & Hopkins & 0.884 & 0.869 & 1.000 / 0.768 \\
        & Silhouette & 0.582 & 0.282 & 1.000 / 0.164 \\
        \hline
        \multirow{3}{*}{30}
        & GNN & \textbf{1.000} & \textbf{1.000} & 1.000 / 1.000 \\
        & Hopkins & 0.901 & 0.890 & 1.000 / 0.802 \\
        & Silhouette & 0.578 & 0.270 & 1.000 / 0.156 \\
        \hline
        \multirow{3}{*}{50}
        & GNN & \textbf{0.999} & \textbf{0.999} & 0.998 / 1.000 \\
        & Hopkins & 0.907 & 0.898 & 1.000 / 0.814 \\
        & Silhouette & 0.576 & 0.264 & 1.000 / 0.152 \\
        \hline
    \end{tabular}
\end{table*}


Other than the 2D and 30D cases discussed earlier, we evaluated the three methods—\mymethod{}, Hopkins Statistic, and Silhouette Score—on dimensions ranging from 2 to 50. Table~\ref{tab:dimension_comparison} summarizes the accuracy, F1 score, and precision/recall for each method. \mymethod{} consistently outperforms the baselines, achieving near-perfect performance as dimensionality increases. In contrast, the Hopkins Statistic shows moderate performance but struggles in higher dimensions, while the Silhouette Score degrades significantly as dimensionality grows.

\subsection{Real-World Test Dataset}
To further evaluate the robustness of \mymethod{}, we conduct experiments on a real-world dataset combined with random noise, simulating scenarios where structured data is faint or sparse. Specifically, we use the MNIST dataset~\cite{deng2012mnist} as the structured data source and generate random noise uniformly distributed within the same range. Two experimental variants are designed to progressively introduce structured data into the noise:

\begin{itemize}
    \item \textbf{Variant 1:} 100\% noise data with an increasing percentage \( p\% \) of structured data (MNIST) added.
    \item \textbf{Variant 2:} The total dataset size remains constant, where \((100-p)\%\) of the data is sampled from noise and \( p\% \) is sampled from structured data.
\end{itemize}

The MNIST dataset is preprocessed to reduce its dimensionality. Each \( 28 \times 28 \) image is first flattened into a 784-dimensional feature vector. We then apply Principal Component Analysis (PCA)~\cite{wold1987principal} to reduce the dimensionality to 50, capturing the most informative features while filtering out noise. After PCA, we randomly sample 200 data points from the MNIST dataset to serve as the structured data. To ensure balance in the experiment, we generate a corresponding noise dataset consisting of 200 uniformly distributed points within the same range as the MNIST data.

In Variant 1, the noise dataset remains constant at 200 points, and we progressively add \( p\% \) of the 200 sampled structured points into the noise. In Variant 2, the total dataset size remains fixed at 200 points, where \((100-p)\%\) are randomly sampled from the noise, and \( p\% \) are drawn from the structured MNIST data. The goal of both variants is to evaluate whether \mymethod{} can detect clustering structures earlier and more reliably compared to baseline methods as the proportion of structured data increases.

We use the same baseline methods described in the synthetic test dataset subsection: one based on the Hopkins Statistic~\cite{hopkins1954new} and the other using K-means clustering with the Silhouette Score~\cite{macqueen1967some, rousseeuw1987silhouettes}. And both baselines are evaluated across multiple thresholds just like in the synthetic data testing.

\begin{figure}[ht]
    \centering
    \begin{subfigure}[b]{0.48\textwidth}
        \centering
        \includegraphics[width=\textwidth]{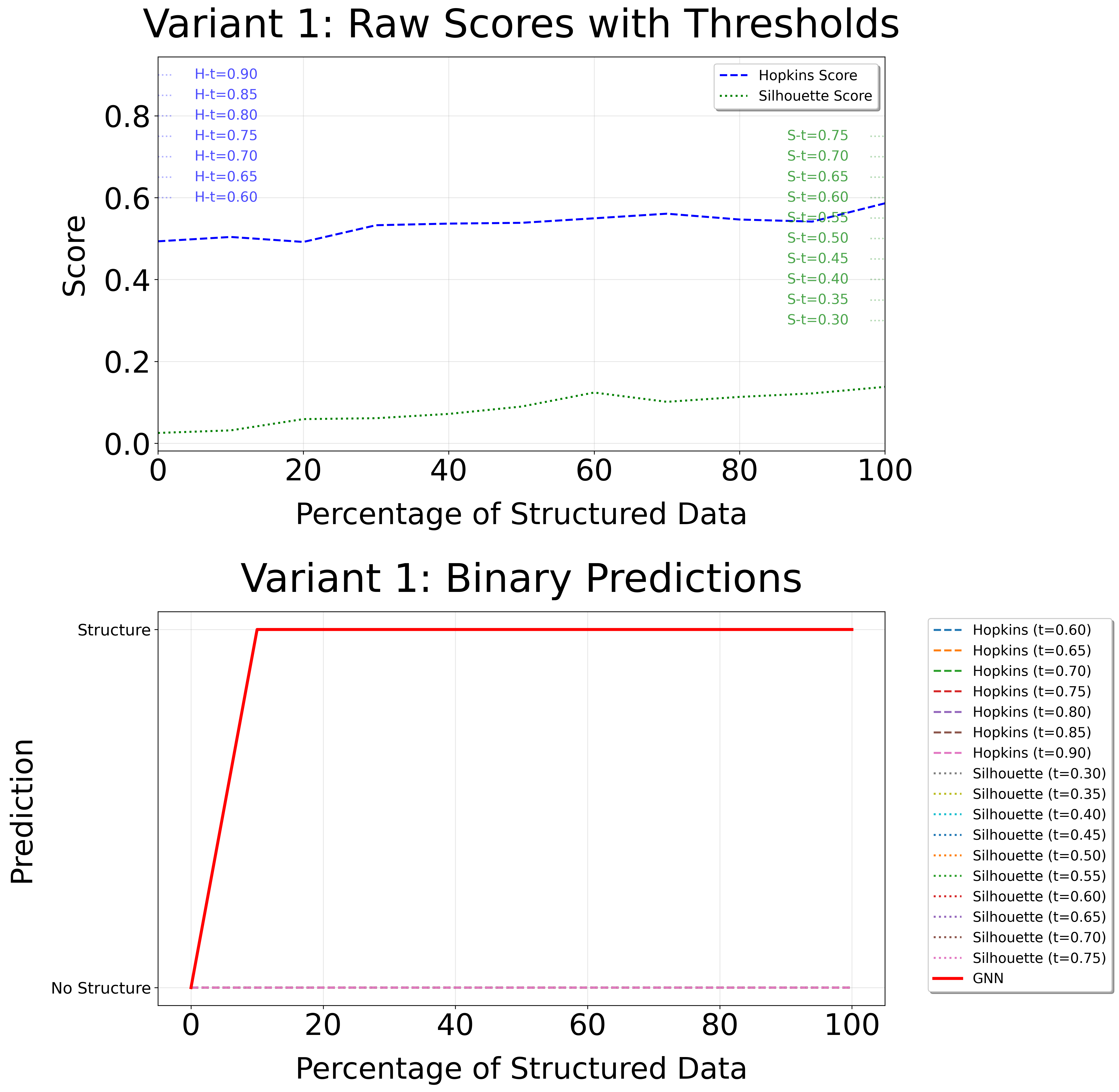}
        \caption{Variant 1: 100\% noise with increasing p\% structured data.}
        \label{fig:mnist_variant1}
    \end{subfigure}
    \hfill
    \begin{subfigure}[b]{0.48\textwidth}
        \centering
        \includegraphics[width=\textwidth]{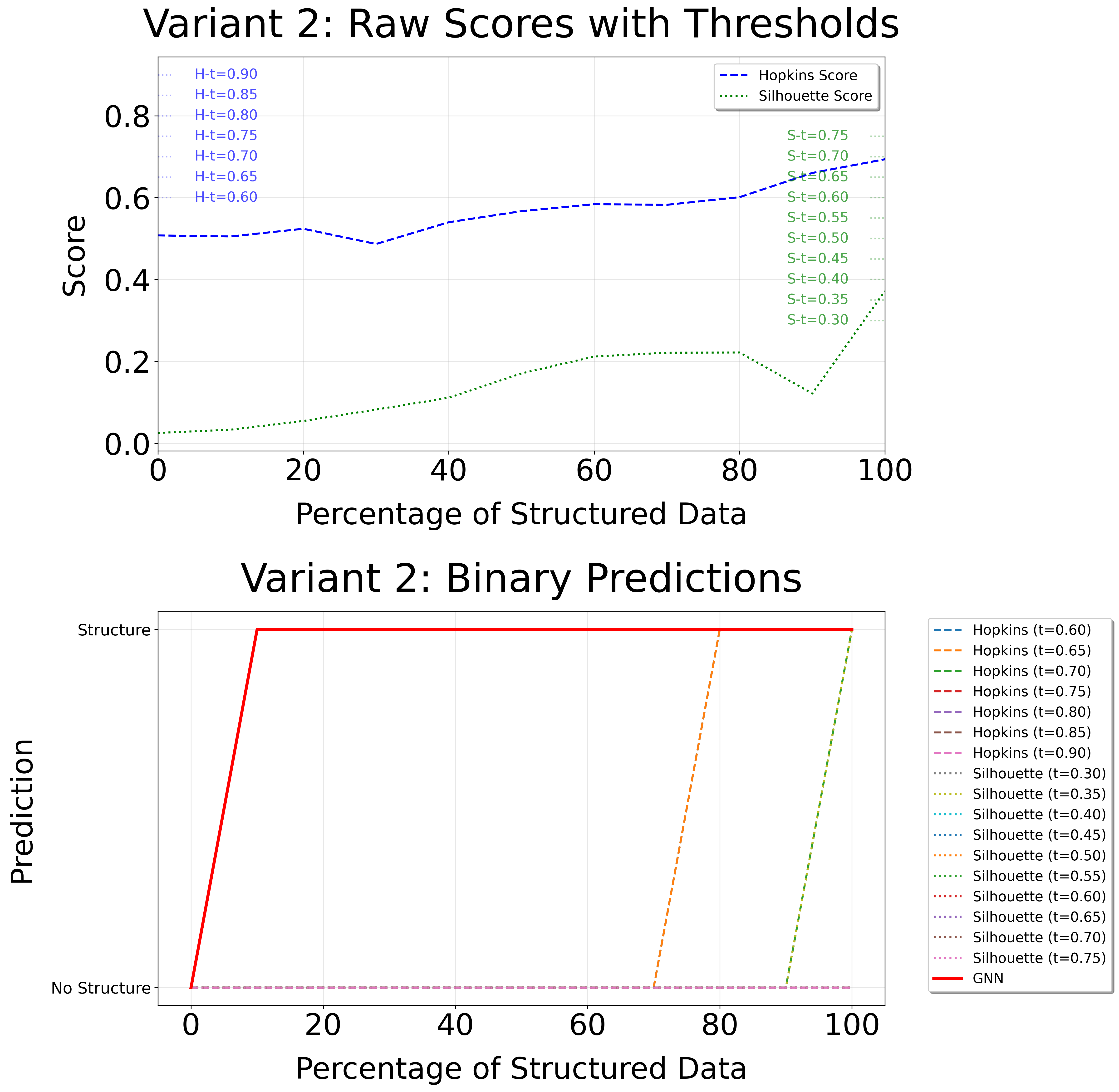}
        \caption{Variant 2: (100-p)\% noise with p\% structured data.}
        \label{fig:mnist_variant2}
    \end{subfigure}
    \caption{Performance comparison of \mymethod{}, Hopkins Statistic, and K-means with Silhouette Score under two experimental variants using the MNIST dataset. The first row in each figure shows the raw scores for the two baseline methods, while the second row presents binary predictions as the percentage of structured data increases.}
    \label{fig:mnist_comparison}
\end{figure}



Figure~\ref{fig:mnist_comparison} shows the experimental results for both variants. In Variant 1 (Figure~\ref{fig:mnist_variant1}), where 100\% noise data is combined with increasing percentages of structured data, the Hopkins Statistic produces relatively flat scores across thresholds and fails to detect structure reliably until the structured data becomes dominant. The Silhouette Score improves marginally at higher percentages but remains inconsistent, especially at lower thresholds. In contrast, \mymethod{} detects clustering structures as early as 10\% structured data, showcasing its superior sensitivity to faint clustering patterns.

In Variant 2 (Figure~\ref{fig:mnist_variant2}), where the noise percentage is progressively reduced, the baselines show delayed improvements, with both methods detecting structure only after 70\% structured data. \mymethod{}, however, consistently identifies clustering structures much earlier, further highlighting its robustness and ability to perform well even when structured data is scarce.

\section{Conclusion}
\label{sec:conclusion}

In this paper, we introduced \mymethod{}, a graph-based method for assessing clustering tendency using Graph Neural Networks (GNNs) trained exclusively on synthetic data. Unlike traditional methods such as the Hopkins Statistic, which require careful threshold tuning and struggle with high-dimensional or noisy data, our approach learns directly from data. By constructing graphs with LSH-based node features and similarity-driven edge features, \mymethod{} effectively captures clustering structures without manual intervention.

Experiments on synthetic datasets showed that \mymethod{} consistently outperformed baseline methods across various dimensions, maintaining high accuracy even as data complexity increased. On real-world datasets mixed with random noise, our method demonstrated superior sensitivity to faint clustering signals, far exceeding baseline performance as the proportion of structured data increased. This highlights \mymethod{}’s robustness and its ability to generalize effectively from synthetic to real-world data.

Our work establishes a learning-based, threshold-free framework for clustering tendency assessment, offering a more reliable solution for modern data analysis. Future directions include exploring alternative graph architectures and validating the method across additional real-world applications.

\section*{Acknowledgments}
Research was sponsored by the Army Research Office and was accomplished under Grant Number W911NF-24-1-0397. The views and conclusions contained in this document are those of the authors and should not be interpreted as representing the official policies, either expressed or implied, of the Army Research Office or the U.S. Government. The U.S. Government is authorized to reproduce and distribute reprints for Government purposes notwithstanding any copyright notation herein.

\bibliographystyle{splncs04}
\bibliography{refs.bib}

\begin{thebibliography}{10}
\providecommand{\url}[1]{\texttt{#1}}
\providecommand{\urlprefix}{URL }
\providecommand{\doi}[1]{https://doi.org/#1}

\bibitem{bezdek2002vat}
Bezdek, J.C., Hathaway, R.J.: Vat: A tool for visual assessment of (cluster) tendency. In: Proceedings of the 2002 International Joint Conference on Neural Networks. IJCNN'02 (Cat. No. 02CH37290). vol.~3, pp. 2225--2230. IEEE (2002)

\bibitem{bhowmick2024dgcluster}
Bhowmick, A., Kosan, M., Huang, Z., Singh, A., Medya, S.: Dgcluster: A neural framework for attributed graph clustering via modularity maximization. In: Proceedings of the AAAI Conference on Artificial Intelligence. vol.~38, pp. 11069--11077 (2024)

\bibitem{deng2012mnist}
Deng, L.: The mnist database of handwritten digit images for machine learning research. IEEE Signal Processing Magazine  \textbf{29}(6),  141--142 (2012)

\bibitem{ghosal2020short}
Ghosal, A., Nandy, A., Das, A.K., Goswami, S., Panday, M.: A short review on different clustering techniques and their applications. Emerging Technology in Modelling and Graphics: Proceedings of IEM Graph 2018 pp. 69--83 (2020)

\bibitem{havens2011efficient}
Havens, T.C., Bezdek, J.C.: An efficient formulation of the improved visual assessment of cluster tendency (ivat) algorithm. IEEE Transactions on Knowledge and Data Engineering  \textbf{24}(5),  813--822 (2011)

\bibitem{hopkins1954new}
Hopkins, B., Skellam, J.G.: A new method for determining the type of distribution of plant individuals. Annals of Botany  \textbf{18}(2),  213--227 (1954)

\bibitem{lloyd1982least}
Lloyd, S.: Least squares quantization in pcm. IEEE transactions on information theory  \textbf{28}(2),  129--137 (1982)

\bibitem{macqueen1967some}
MacQueen, J., et~al.: Some methods for classification and analysis of multivariate observations. In: Proceedings of the fifth Berkeley symposium on mathematical statistics and probability. vol.~1, pp. 281--297. Oakland, CA, USA (1967)

\bibitem{pagadala2024havat}
Pagadala, K.M., Rathore, P.: Havat: Automatic cluster structure assessment in unlabeled data. In: Proceedings of the 7th Joint International Conference on Data Science \& Management of Data (11th ACM IKDD CODS and 29th COMAD). pp. 45--53 (2024)

\bibitem{rousseeuw1987silhouettes}
Rousseeuw, P.J.: Silhouettes: a graphical aid to the interpretation and validation of cluster analysis. Journal of computational and applied mathematics  \textbf{20},  53--65 (1987)

\bibitem{shiao2024frappe}
Shiao, W., Papalexakis, E.E.: Frappe: fast rank approximation with explainable features for tensors. Data Mining and Knowledge Discovery  \textbf{38}(6),  4217--4232 (2024)

\bibitem{tsitsulin2023graph}
Tsitsulin, A., Palowitch, J., Perozzi, B., M{\"u}ller, E.: Graph clustering with graph neural networks. Journal of Machine Learning Research  \textbf{24}(127),  1--21 (2023)

\bibitem{wang2010ivat}
Wang, L., Nguyen, U.T., Bezdek, J.C., Leckie, C.A., Ramamohanarao, K.: ivat and avat: enhanced visual analysis for cluster tendency assessment. In: Pacific-Asia Conference on Knowledge Discovery and Data Mining. pp. 16--27. Springer (2010)

\bibitem{wold1987principal}
Wold, S., Esbensen, K., Geladi, P.: Principal component analysis. Chemometrics and intelligent laboratory systems  \textbf{2}(1-3),  37--52 (1987)

\bibitem{yuan2023real}
Yuan, J., Zhang, J., Sun, S., Torr, P., Zhao, B.: Real-fake: Effective training data synthesis through distribution matching. arXiv preprint arXiv:2310.10402  (2023)

\bibitem{zhang2022non}
Zhang, H., Shi, J., Zhang, R., Li, X.: Non-graph data clustering via {O(n)} bipartite graph convolution. IEEE Transactions on Pattern Analysis and Machine Intelligence  \textbf{45}(7),  8729--8742 (2022)

\end{thebibliography}

\appendix
\section{Appendix: Analysis of Edge Features and Neighbor Connections}  
\begin{figure}[ht]
    \centering
    \includegraphics[width=0.95\textwidth]{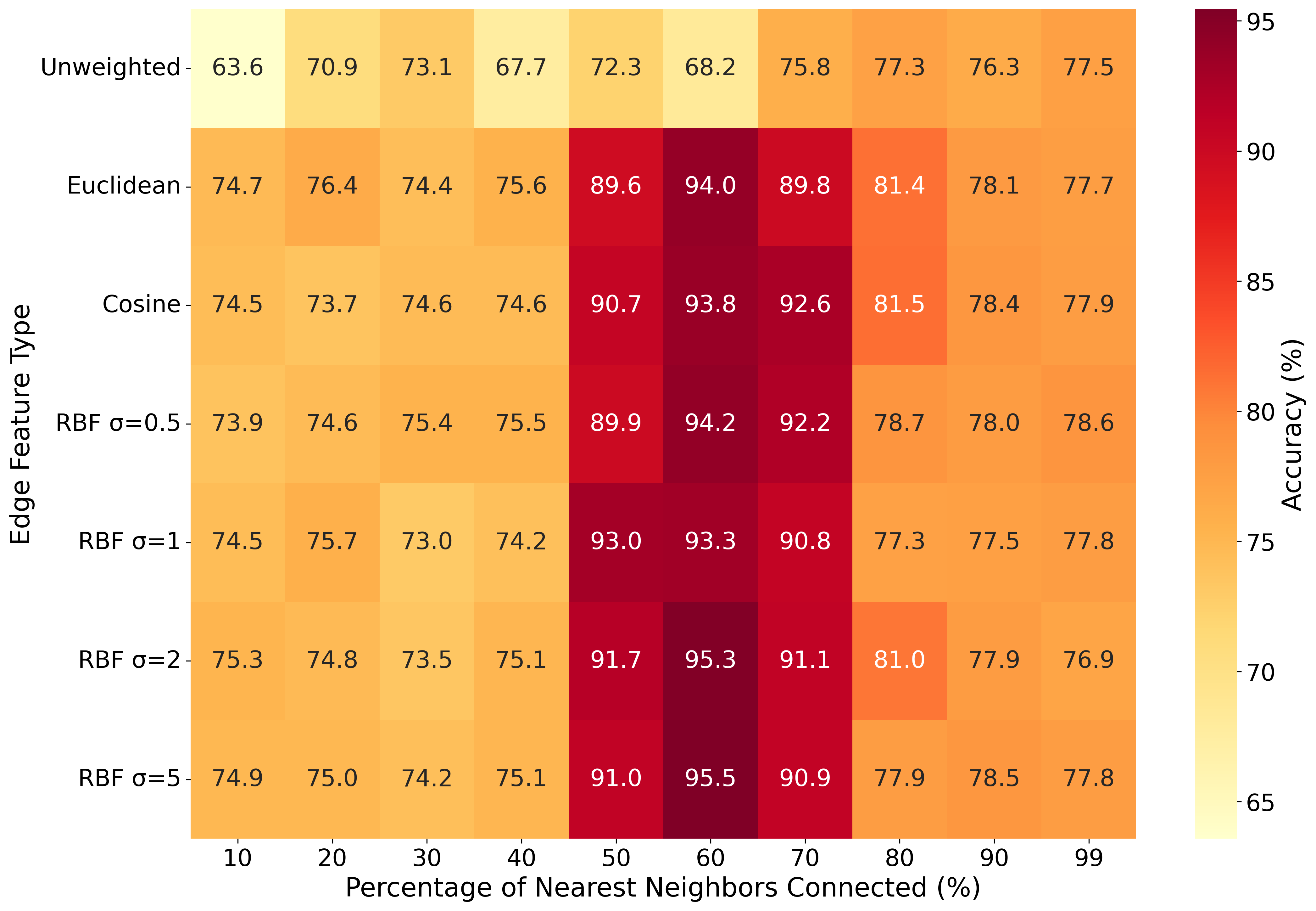}
    \caption{Heatmap of testing accuracy for different edge feature strategies and percentages of nearest neighbors connected. The RBF kernel with moderate $\sigma$ values (2 or 5) and 50\%--60\% neighbor connections achieves the highest accuracy.}  
    \label{fig:accuracy_heatmap}
\end{figure}

To determine the optimal combination of edge feature construction and the percentage of nearest neighbors connected, we conducted a grid search experiment on 2D synthetic data. The results are visualized as a heatmap in Figure~\ref{fig:accuracy_heatmap}, showing testing accuracy across different configurations.

The horizontal axis represents the percentage of nearest neighbors connected, ranging from 10\% to 99\% in increments of 10\%. The vertical axis corresponds to different edge feature strategies, including unweighted edges, Euclidean distance, cosine similarity, and RBF kernels (with varying $\sigma$ values).  

From the heatmap, we observe that accuracy peaks when 50\%--60\% of the nearest neighbors are connected, regardless of the edge feature strategy. RBF kernels with $\sigma = 2$ and $\sigma = 5$ consistently achieve the highest accuracy, particularly with 50\%--70\% connectivity. Euclidean distance and cosine similarity perform well but fall slightly short of the RBF-based methods. In contrast, unweighted edges show significantly lower accuracy, emphasizing the importance of meaningful edge features.

These findings informed the choice of RBF-based edge features (with $\sigma = 2$ or $\sigma = 5$) and a moderate neighbor connection percentage (50\%--60\%) in our main experiments.  

\end{document}